\documentclass[10pt,twocolumn,letterpaper]{article}

\usepackage{cvpr}
\usepackage{times}
\usepackage{epsfig}
\usepackage{graphicx}
\usepackage{amsmath,amssymb,graphicx,subfigure,multirow,float,makecell}
\usepackage{booktabs}
\usepackage{listings}

\usepackage[breaklinks=true,bookmarks=false]{hyperref}
\cvprfinalcopy 

\def\cvprPaperID{3916}

\begin{document}
\title{Attention-based Dropout Layer for Weakly Supervised Object Localization}
\author{Junsuk Choe, and Hyunjung Shim\thanks{Corresponding author.}\\
School of Integrated Technology, Yonsei University, South Korea\\
{\tt\small $\lbrace$junsukchoe, kateshim$\rbrace$@yonsei.ac.kr}
}

\maketitle

\begin{abstract}
Weakly Supervised Object Localization (WSOL) techniques learn the object location only using image-level labels, without location annotations. A common limitation for these techniques is that they cover only the most discriminative part of the object, not the entire object. To address this problem, we propose an Attention-based Dropout Layer (ADL), which utilizes the self-attention mechanism to process the feature maps of the model. The proposed method is composed of two key components: 1) hiding the most discriminative part from the model for capturing the integral extent of object, and 2) highlighting the informative region for improving the recognition power of the model. Based on extensive experiments, we demonstrate that the proposed method is effective to improve the accuracy of WSOL, achieving a new state-of-the-art localization accuracy in CUB-200-2011 dataset. We also show that the proposed method is much more efficient in terms of both parameter and computation overheads than existing techniques.
\end{abstract}\vspace{-4mm}

\section{Introduction}
\label{sec:introduction}

Weakly Supervised Object Localization (WSOL) aims to identify the location of the object in a scene only using image-level labels, not location annotations. Existing approaches mine and track discriminative features of each class for object detection \cite{wang2014weakly, song2014learning, song2014weakly, cinbis2014multi, wang2014weakly, oquab2015object, liang2015towards, teh2016attention, li2016weakly, bilen2016weakly, sun2016pronet, kantorov2016context, zhou2016learning, durand2017wildcat, dong2017dual, diba2017weakly, wei2017selective, jie2017deep, zhu2017soft, shi2017weakly, selvaraju2017grad, zhang2018top, zhang2018ml, gao2017cwsl,  dong2018few} and segmentation \cite{simonyan2013deep, pathak2015constrained, kolesnikov2016seed, khoreva2016weakly, oh2017exploiting, wei2018revisiting, wei2017stc}. Because the discriminative power of each object part is different from another, these techniques tend to identify only the most discriminative part of the target object, incapable of covering entire extent of the object. For example, in the case of a person, the face may be more discriminative than the body which appearance changes dramatically due to clothing. In this case, existing WSOL techniques can localize only the face, not the entire region.

This problem can be critical in object localization. Specifically, Class Activation Mappings (CAM) \cite{zhou2016learning} utilize Convolutional Neural Networks (CNN) classifier for learning the discriminative features. The key idea is that the classifier with a reasonable accuracy should observe the object region to decide the class label. In other words, the discriminative features should co-occur with the object region. From this idea, they perform localization by tracking spatial distribution of feature responses. Unfortunately, the classifiers tend to focus only on the most discriminative features to increase their classification accuracy. Therefore, the spatial distribution of feature responses also tends to cover only the most discriminative part of the object, which leads to localization accuracy degradation.  

\begin{figure*}[t]
\centering
\includegraphics[height=4.25cm]{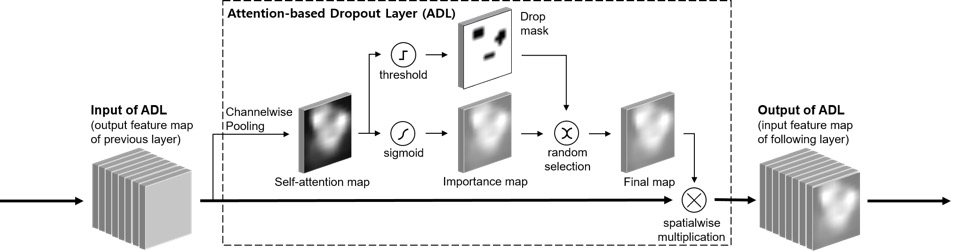}\vspace{2mm}
\caption{ADL block diagram. The self-attention map is generated by channelwise average pooling of the input feature map. Based on the self-attention map, we produce a drop mask using thresholding and an importance map using a sigmoid activation, respectively. The drop mask and the importance map are selected stochastically at each iteration and applied to the input feature map. Please note that this figure illustrates the case when the importance map is selected.}
\label{fig:block}
\end{figure*}

Recently, various techniques \cite{wei2017object, singh2017hide, kim2017two, zhang2018adversarial, li2018tell, wei2018revisiting, wei2018ts2c, zhang2018self} have been proposed to address this issue. Most of them \cite{wei2017object, kim2017two, singh2017hide, zhang2018adversarial, li2018tell} erased the most discriminative region on the input image or feature map by zeroing that region during the training phase. These techniques are similar to the dropout \cite{srivastava2014dropout} in that they deactivate specific nodes of the feature map by setting them zero during the training phase. This prevents the model from relying solely on the most discriminative part for classification, instead encourages it to learn the less discriminative part as well. To achieve this goal, Hide-and-Seek (HaS) \cite{singh2017hide} divides the input image into grid-like patches and randomly selects the patches to erase. While the random selection is simple and fast, it cannot effectively erase the most discriminative part. 

For effectively removing only the most discriminative part, several techniques \cite{wei2017object, kim2017two, zhang2018adversarial, li2018tell} have been proposed. These techniques re-train the model multiple times \cite{wei2017object, kim2017two}, use additional classifiers \cite{kim2017two, zhang2018adversarial}, or perform two forward-backward propagations per one iteration \cite{li2018tell} for finding the most discriminative part. Consequently, huge additional computing resources are required to eliminate the most discriminative part effectively.

From previous methods, we conclude that the idea of erasing only the most discriminative part is effective to capture the full extent of object. However, existing methods require substantial computing resources to remove the most discriminative part accurately. Our goal is to erase the most discriminative part in an effective and efficient way. To this end, we propose an Attention-based Dropout Layer (ADL), a lightweight yet powerful method which utilizes self-attention mechanism to remove the most discriminative part of the target object.

Specifically, a self-attention map is obtained by performing channelwise average pooling on the input feature map. Based on the self-attention map, we produce two key components of ADL, a \emph{drop mask} and an \emph{importance map}. The drop mask is used to hide the most discriminative part during training. This induces the model to learn the less discriminative part as well. We obtain this drop mask by thresholding the self-attention map. The importance map is used to highlight informative region for improving the classification power of the model. Owing to the importance map, the more accurate self-attention map can be produced. The importance map is computed by applying sigmoid activation to the self-attention map. During training, either one of the drop mask or importance map is stochastically selected at each iteration, and then the selected one is applied to the input feature map by spatialwise multiplication. Figure \ref{fig:block} shows the block diagram of the proposed method.

Compared to existing WSOL techniques, the proposed method is much more efficient in terms of both computation and parameter overheads. This is because we can find and erase the most discriminative region by a single forward-backward propagation in a single model. In addition, regardless of the model architecture, ADL can be easily applied to convolutional feature maps of the model to improve the localization accuracy. Compared to existing self-attention techniques \cite{wang2017residual, hu2018squeeze, park2018bam, woo2018cbam}, the proposed method is greatly lightweight because there are no additional trainable parameters for extracting self-attention map. 

The proposed method is lightweight and efficient, and also report excellent accuracy. Quantitatively, the proposed method achieves superior accuracy, more than 15 percentage points of accuracy improvement, over the existing state-of-the-art techniques \cite{zhang2018adversarial, zhang2018self} on CUB-200-2011 dataset \cite{wah2011cub}, and comparable accuracy to the current state-of-the-art technique \cite{zhang2018self} on ImageNet-1k dataset \cite{russakovsky2015imagenet}. We also observe consistent results in qualitative evaluation; the model with ADL learns the less discriminative part better than the vanilla model \cite{zhou2016learning}. 

\vspace{2mm}
\section{Related Work}
\label{sec:relatedworks}

\noindent\textbf{Dropout.} Dropout \cite{srivastava2014dropout} is a regularization technique to alleviate overfitting in neural networks. Specifically, dropout discards information by randomly zeroing each hidden node of the neural network during the training phase. In this way, the network can enjoy the ensemble effect of small subnetworks, thus achieving a good regularization effect. However, unlike fully connected layers, applying dropout to the convolutional feature map is not effective. One of the reasons is that spatially adjacent pixels are strongly correlated on the convolutional feature map; they share redundant contextual information. Hence, the conventional pixel-based dropout cannot completely discard the information on the convolutional feature map \cite{tompson2015efficient}.

In order to apply dropout to the convolutional feature map, Tompson \textit{et al.} \cite{tompson2015efficient} proposed SpatialDropout that randomly drops partial channels of a feature map, rather than dropping each pixel. Based on this channel-based dropout, the problem of pixel-level dropout can be resolved. The proposed method differs from SpatialDropout in that we drop only strongly activated regions, rather than dropping entire region of channel. Owing to this region-based dropout, we could also bypass the problem of pixel-level dropout. Meanwhile, Park and Kwak \cite{park2016analysis} proposed MaxDrop, which drops the maximally activated pixel through channelwise or spatialwise on the feature map. Similar to MaxDrop, the proposed method drops strongly activated part. However, we differ from MaxDrop in that we use attention mechanism to find the maximally activated part. In addition, the proposed method does not drop the maximally activated \emph{pixel}, but maximally activated \emph{region}.

\noindent\textbf{Attention mechanism.} Humans selectively use an important part of the data to make a decision \cite{corbetta2002control, itti1998model}. Similarly, when a query comes in, the artificial model does not process all the data equally, but focuses only on the important data. This process is called \emph{attention mechanism} and is actively used in various fields such as machine translation \cite{vaswani2017attention}, image captioning \cite{xu2015show}, image inpainting \cite{yu2018generative, liu2018image}, transfer learning \cite{zagoruyko2017paying}, visual question answering \cite{zhu2016visual7w}, and generative model \cite{parmar2018image, zhang2018self}. When the query is input itself, such attention is specifically called \emph{self-attention}, which is effective to learn the meaningful representation for conducting the given task. For example, in the case of the classification task, the self-attention map appears in a form that emphasizes informative features for classification (\textit{e.g.}, the most discriminative part of the target object).

Recently, various methods \cite{wang2017residual, wang2017non, hu2018squeeze, park2018bam, woo2018cbam} utilize the self-attention mechanism to enhance the accuracy of the CNN classification model. Residual Attention Networks (RAN) \cite{wang2017residual} has improved the accuracy of the classification model using 3D self-attention map. However, the parameter overheads are very large because the raw feature map without any compression is used for attention extraction. Squeeze-and-Excitation (SE) \cite{hu2018squeeze} increases the accuracy of the classification model using only 1D channel self-attention map. For extracting the self-attention map, the feature map is first compressed using Global Average Pooling (GAP) and then passed through 2-layer MLP. In this way, SE can significantly reduce the parameter overheads for attention extraction compared to RAN. However, the parameter overheads are still not negligible (\textit{e.g.}, 10\% on ResNet50 \cite{he2016deep}). 

Bottleneck Attention Module (BAM) \cite{park2018bam} and Convolutional Block Attention Module (CBAM) \cite{woo2018cbam} increase the accuracy of the classifier by utilizing both 1D channel and 2D spatial self-attention maps. They compute the spatial self-attention map using auxiliary convolutional layer(s). The computed self-attention map is applied to the input feature map for rewarding the informative region. Likewise, the proposed method uses the \textit{importance map} for rewarding the informative region. However, the key difference from them is that we stochastically penalize that region using the \textit{drop mask}. Also, unlike these techniques, we do not require additional trainable parameters for extracting the self-attention map.

\section{ADL: Attention-based Dropout Layer}
\label{sec:adl}

\begin{figure*}[t]
\centering
\includegraphics[width=1.7\columnwidth]{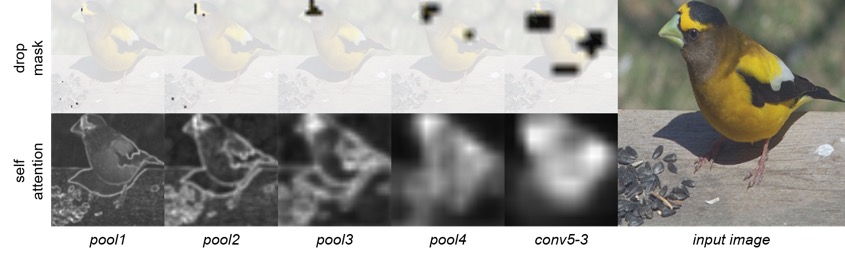}

\caption{Drop mask and self-attention map at each layer of VGG-GAP \cite{zhou2016learning}. At lower-level layers, the self-attention maps include general features, while class-specific features are included in the self-attention maps at higher-level layers. The drop masks also erase most discriminative part more effectively at higher-level layers. Please note that the drop mask is overlaid with input image for better visualization. Because the importance map has a distribution very similar to that of the self-attention map, we do not visualize it.}
\label{fig:adlviz}
\vspace{-2mm}
\end{figure*}

In this section, we present details of the proposed method, Attention-based Dropout Layer (ADL). ADL is applied on each feature map of classification model, and induces the model to learn the entire region of the object. ADL generates a self-attention map from input feature map, and produces a drop mask and an importance map. Although both components are computed from self-attention map, they play the opposite role. The drop mask penalizes the most discriminative part for inducing the model to cover the integral extent of the object. Meanwhile, the importance map rewards the most discriminative part for increasing the classification power of the model. During training, the drop mask or importance map is stochastically selected for each iteration. Then, the selected one is applied to the input feature map. By applying each component stochastically, we can enjoy their advantages simultaneously. ADL has two main hyperparameters: \textit{drop\_rate} and $\gamma$. The \textit{drop\_rate} indicates how frequently the drop mask is applied, and the $\gamma$ controls the size of the region to be dropped.  The example of each component is visualized in Figure \ref{fig:adlviz}.

Specifically, the input of ADL is a convolutional feature map $ \textbf{F}\subseteq \textbf{R}^{H\times W \times C} $. Note that $C$ is the number of channels, $H$ and $W$ are height and width, respectively. For simplicity, we omit the mini-batch dimension in this notation. We generate a self-attention map $ \textbf{M}_{att} \subseteq \mathrm{\textbf{R}}^{H\times W} $   by compressing $\textbf{F}$ using channelwise average pooling. Because the model is trained for classification, the intensity of each pixel in the self-attention map is proportional to the discriminative power. In this way, we can approximate the spatial distribution of the most discriminative part efficiently. 

To obtain the drop mask, we first set a drop threshold by prefixed ratio $\gamma$ of maximum intensity of the self-attention map. Then, we produce the drop mask $ \textbf{M}_{drop} \subseteq \mathrm{\textbf{R}}^{H\times W} $ by setting each pixel to $0$ if it is larger than drop threshold, and $1$ if it is smaller. That is, the drop mask has $0$ for the most discriminative region and $1$ for otherwise. Note that the size of region to be dropped increases as $\gamma$ decreases, and vice versa. The drop mask is applied to the input feature map by spatialwise multiplication. In this way, we can hide the most discriminative part from the model; we encourage the model to learn the less discriminative part for classification but meaningful region for localization. However, if the drop mask is applied at every iteration, the most discriminative part is never observed during the training phase. As a result, the classification accuracy of the model is significantly decreased, which adversely affects the localization accuracy. To remedy this, we stochastically apply the drop mask according to \textit{drop\_rate}. When the drop mask is not applied, the importance map is applied instead. We generate the importance map $ \textbf{M}_{imp} \subseteq \mathrm{\textbf{R}}^{H\times W} $ from the self-attention map by applying sigmoid activation. That is, the intensity of each pixel in the importance map is close to $1$ for the most discriminative region, and close to $0$ for the less discriminative region. Like the drop mask, the importance map is applied to the input feature map by spatialwise multiplication. In this way, we can improve the classification accuracy of the model.  

The proposed method is applied independently to each convolutional feature map. Therefore, it can be easily plugged into multiple feature maps of existing classification models for improving localization accuracy. In addition, it does not require any trainable parameters. That means, there is no parameter overheads even when applied to multiple feature maps at the same time. Furthermore, with ADL, the most discriminative region can be identified and erased efficiently, without auxiliary classifiers \cite{kim2017two, zhang2018adversarial}, re-training \cite{wei2017object}, or additional forward-backward propagation \cite{li2018tell}. 

ADL is an auxiliary module which is applied only during training. During the testing phase, ADL is deactivated. That is, our testing phase is identical to that of vanilla model. Therefore, the object localization can be performed using various heatmap extraction methods \cite{zhou2016learning, selvaraju2017grad, zhang2018adversarial, li2018tell} without bells and whistles. Note that we do not compensate the different distributions between training and testing, as other dropout-based WSOL techniques \cite{kim2017two, singh2017hide, zhang2018adversarial}. 

\noindent \textbf{Relation with other attention extraction methods.} Our extraction method does not require trainable parameters, much lightweight compared to existing methods \cite{wang2017residual, hu2018squeeze, park2018bam, woo2018cbam}. Thus, one might wonder how our method can produce semantically meaningful results despite its simplicity. 

Recently, Zagorukyo and Komodakis \cite{zagoruyko2017paying} showed that the informative region for transfer learning can be identified by applying the channelwise pooling to the convolutional feature map. That is, the self-attention map for transfer learning is obtained by the channelwise pooling. Inspired by this, CBAM \cite{woo2018cbam} utilized the self-attention map to improve the classification accuracy. Specifically, they refine the map using auxiliary convolutional layer and sigmoid activation. This refined self-attention map is applied to input feature map by spatialwise multiplication. In this way, the auxiliary convolutional layers are trained to refine the self-attention map for improving the classification accuracy. 

However, from the empirical study, we observe that the self-attention map may not need to be refined by auxiliary layers. We conjecture that it is because existing convolutional layers in CNN model are sufficiently powerful to produce meaningful self-attention map. Hence, after computing the self-attention map by channelwise average pooling, we normalize this map using sigmoid activation and then multiply it to input feature map. Then, the gradient from the loss function updates existing convolutional layers so that the resultant self-attention map is useful for improving classification accuracy. For example, if the self-attention map fails to highlight the object region, this may degrade the classification accuracy. Hence, existing convolutional layers are trained to produce more accurate self-attention map. This is equivalent to assigning the role of the auxiliary convolutional layer used in CBAM to the existing convolutional layers in the model. Note that the similar principle was introduced by Lin \textit{et al.} \cite{lin2014network}; they replace the fully connected layers of CNN classifier with the GAP layer. 

The improvement of classification accuracy of our attention method may not be as great as that of CBAM. However, our method is much more efficient and can produce sufficiently meaningful results for our application. This is shown in our experimental results; our self-attention map is effective to increase classification accuracy and identify the most discriminative part of the target object.

\begin{figure*}[t!]
    \noindent
    \begin{center}
        \includegraphics[width=2.0\columnwidth]{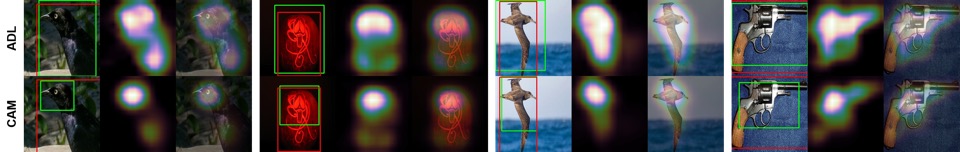}
    \end{center}
    \caption{Qualitative evaluation results of VGG-GAP \cite{zhou2016learning} on CUB-200-2011 and ImageNet-1k. The left image in each figure is input image. The red bounding box is ground truth, while the green bounding box is estimates. The middle image is heatmap and the right image shows the overlap between the input image and the heatmap. We also compared our method and the vanilla model side by side.}
\label{fig:qualitative}
\vspace{-3mm}
\end{figure*}

\noindent \textbf{Relation between drop mask and importance map.} In our model, the drop mask penalizes the most discriminative part, while the importance map rewards the most discriminative part. One might consider the drop mask and importance map are mutually exclusive. However, our experimental results support that they are not mutually exclusive. We believe that it is because the drop mask can be accurately produced by the importance map. Specifically, as the importance map improves the classification accuracy, the more accurate self-attention map can be produced. Consequently, the drop mask can more effectively erase the most discriminative region of the object.

\noindent \textbf{Relation between classification and localization.} Previous study \cite{singh2017hide} has reported that the classification accuracy is compromised while the localization accuracy is increased. They conjecture that this is caused by the usage of a drop mask. Because we also use a drop mask to erase the most discriminative part, such a trade-off relationship between the accuracy of localization and that of classification is consistently observed in our experiments. However, the proposed method can boost the classification power using the importance map, thus the accuracy degradation of classification is not as significant as other techniques.

\noindent \textbf{Relation with the current state-of-the-arts.} The current state-of-the-art techniques for WSOL are Adversarial Complementary Learning (ACoL) \cite{zhang2018adversarial} and Self-Produced Guidance (SPG) \cite{zhang2018self}. ACoL adds two auxiliary classifiers in parallel to the backbone feature extractor for finding the most discriminative part of the target object. The proposed method differs from ACoL in that we can find the most discriminative part without the additional classifier, which is much more efficient. Most recently, SPG has been proposed, a new WSOL technique that utilizes spatial distribution of the object and background. The classifier can learn the integral extent of the object using that distribution as auxiliary supervision. The proposed method differs from SPG in that SPG does not erase the most discriminative part of the object. In addition, SPG requires substantial computing resources for improving the localization accuracy. 

\section{Experimental Results}
\label{sec:experiments}

\noindent\textbf{Dataset.} We evaluate the performance of the proposed method in CUB-200-2011 \cite{wah2011cub} and ImageNet-1k \cite{russakovsky2015imagenet}, respectively. The ImageNet-1k is a large-scale dataset with 1,000 different classes, consisting of approximately 1.3 million training images and 50,000 validation images. For this dataset, we train the model with the training set and evaluate the performance with the validation set. 

The CUB-200-2011 includes 200 species of birds, consisting of 5,994 training images and 5,794 testing images. For this dataset, we train the model with the training set and evaluate the performance with the testing set. The intra-class variation of CUB-200-2011 is smaller than that of ImageNet-1k, because all classes of this dataset belong to \emph{birds}. In this case, the extent of the most discriminative region might be quite small. For example, in \emph{Common Raven} and \emph{White-necked Raven}, there is no difference in appearance except the color of the neck. That is, the most discriminative part is the neck, which is very small compared to the entire area of the bird. Consequently, although CUB-200-2011 is not a large-scale dataset such as ImageNet-1k, this is a particularly challenging dataset to conduct WSOL.

\noindent \textbf{Implementation details.} We use VGG \cite{simonyan2014very}, ResNet \cite{he2016deep}, MobileNetV1 \cite{howard2017mobilenets}, and InceptionV3 \cite{szegedy2016rethinking} as backbone networks. Note that we replace the last pooling layer and two fully connected layers of VGG16 with a GAP layer, according to \cite{zhou2016learning}. We also use the customized InceptionV3 as a backbone, following the SPG \cite{zhang2018self}. We plug SE block \cite{hu2018squeeze} into ResNet50 for demonstrating the compatibility of ADL with other self-attention methods. For ResNet and MobileNetV1, we set the stride of last strided convolution to 1 for enlarging the spatial resolution of heatmap to 14$\times$14.

ADL is plugged in each feature map of the CNN model in a sequential way; the output of ADL is the input of the next layer. We use a pre-trained model which is trained with ImageNet-1k dataset \cite{russakovsky2015imagenet}, and then fine-tune the network. We extract the heatmap from classification model using CAM \cite{zhou2016learning}. Also, the bounding box is extracted from the heatmap using the same method as presented in \cite{zhou2016learning}. We implement the models using Tensorpack \cite{wu2016tensorpack} on Tensorflow \cite{abadi2016tensorflow}, and train them using NVIDIA Titan Xp GPU.

Based on extensive ablation studies, we find that it is optimal to apply ADL to intermediate and higher-level layers of the network. Especially, for the intermediate layer, it is preferable to apply it to bottleneck part (\textit{e.g.}, pooling layer or strided convolution). We set the \textit{drop\_rate} as 75\%. For the drop threshold, we set $\gamma$ to 80\% for VGG-GAP and InceptionV3, 90\% for ResNet, and 95\% for MobileNetV1. However, the hyperparameters mentioned here are only the recommended settings. Note that the localization accuracy can be further improved when the optimal setting is used.

\noindent\textbf{Metrics.} We use three evaluation metrics as \cite{singh2017hide}: Top-1 classification accuracy (\textit{Top-1 Clas}), Localization accuracy with known ground-truth class (\textit{GT-known Loc}), and Top-1 localization accuracy (\textit{Top-1 Loc}). \textit{Top-1 Clas} determines that the answer is correct when the estimated class is equal to the ground truth class. \textit{GT-known Loc} judges the answer as correct when the intersection over union (IoU) between the ground truth bounding box and estimated box for the ground truth class is 50\% or more. Lastly, \textit{Top-1 Loc} considers the answer as correct when  both \textit{Top-1 Clas} and \textit{GT-known Loc} are correct. Please note that it is considered to be the most appropriate to use \textit{Top-1 Loc} for evaluating overall localization performance, according to \cite{russakovsky2015imagenet}. 

\begin{table}[]
\small
\centering
\begin{tabular}{@{}ccccc@{}}
\toprule
\begin{tabular}[c]{@{}c@{}}Drop\\ mask (\%)\end{tabular} & \begin{tabular}[c]{@{}c@{}}Importance\\ map (\%)\end{tabular} & \begin{tabular}[c]{@{}c@{}}GT-known\\ Acc (\%)\end{tabular} & \begin{tabular}[c]{@{}c@{}}Top-1\\ Clas (\%)\end{tabular} & \begin{tabular}[c]{@{}c@{}}Top-1\\ Loc (\%)\end{tabular} \\ \midrule
100                                                      & \multicolumn{1}{c|}{0}                                        & 72.43                                                       & \multicolumn{1}{c|}{57.37}                                & 44.11                                                    \\
75                                                       & \multicolumn{1}{c|}{25}                                       & \textbf{74.78}                                              & \multicolumn{1}{c|}{62.25}                                & \textbf{49.69}                                           \\
50                                                       & \multicolumn{1}{c|}{50}                                       & 71.51                                                       & \multicolumn{1}{c|}{64.93}                                & 49.33                                                    \\
25                                                       & \multicolumn{1}{c|}{75}                                       & 67.29                                                       & \multicolumn{1}{c|}{\textit{68.99}}                       & 47.98                                                    \\
0                                                        & \multicolumn{1}{c|}{100}                                      & 47.51                                                       & \multicolumn{1}{c|}{67.78}                                & 32.24                                                    \\ \midrule
N/A                                                      & \multicolumn{1}{c|}{N/A}                                      & 51.09                                                       & \multicolumn{1}{c|}{67.55}                                & 34.41                                                    \\ \midrule
75                                                       & \multicolumn{1}{c|}{N/A}                                      & 73.23                                                       & \multicolumn{1}{c|}{61.55}                                & 47.67                                                    \\
N/A                                                      & \multicolumn{1}{c|}{25}                                       & 50.62                                                       & \multicolumn{1}{c|}{\textit{68.50}}                       & 33.91                                                    \\
75                                                       & \multicolumn{1}{c|}{25}                                       & \textbf{74.78}                                              & \multicolumn{1}{c|}{62.25}                                & \textbf{49.69}                                           \\ \bottomrule
\end{tabular}\vspace{2mm}
\caption{Upper: Accuracy according to \textit{drop\_rate}. Middle: Baseline accuracy. Lower: Accuracy when each component has been deactivated. Bold text refers the best localization accuracy, while \textit{italic text} refers the best classification accuracy. N/A indicates that ADL outputs the raw input feature map instead of applying drop mask or importance map.}
\label{tab:tab1}

\end{table}\noindent 
\vspace{-2mm}
\subsection{Ablation Study}

In this subsection, we utilize pre-trained VGG-GAP \cite{simonyan2014very, zhou2016learning} as a backbone network. For training, we plug ADLs in all the pooling layers and the \emph{conv5-3} layer, and then fine-tune the model using CUB-200-2011 dataset.

First, we visualize the self-attention map and drop mask in Figure \ref{fig:adlviz}. We observe that the self-attention maps of lower-level layers (\textit{i.e.}, \emph{pool1} and \emph{pool2}) contain class-agnostic general features. Meanwhile, the self-attention maps of higher-level layers (\textit{i.e.}, \emph{pool4} and \emph{conv5-3}) contain the class-specific features. We also observe that the drop masks from higher-level layers erase the most discriminative part more accurately than those from lower-level layers.

Next, we investigate the effect of \textit{drop\_rate} on accuracy. The upper part of Table \ref{tab:tab1} reports the results. From these results, we observe that the best localization accuracy can be achieved when the \textit{drop\_rate} is 75\%. Meanwhile, when the drop mask is applied at every iteration (\textit{i.e.}, \textit{drop\_rate} 100\%), the classification (\textit{Top-1 Clas}) and localization (\textit{Top-1 Loc}) accuracy are greatly reduced. This is because, as mentioned in Section \ref{sec:adl}, the model never observe the most discriminative part. As a result, the classification power of the model decreases significantly, which adversely influences localization accuracy.  Given that accuracy degradation in \textit{GT-known Acc} is relatively less than that of \textit{Top-1 Loc} and that of \textit{Top-1 Clas}, we can conclude that this is the result of the classification accuracy degradation.

We observe that the classification accuracy increases as the \textit{drop\_rate} decreases. However, when the \textit{drop\_rate} becomes too low (\textit{drop\_rate} from 25\% to 0\%), the classification accuracy decreases again (from 68.99\% to 67.78\%). We believe this is caused by overfitting. The drop mask is a dropout-based technique and its rationale is similar to MaxDrop \cite{park2016analysis}. Thus, the drop mask with proper \textit{drop\_rate} may prevent overfitting, increasing the classification accuracy. We consider that the analysis of regularization effect of the drop mask is beyond the scope of this paper. Yet, we plan to analyze this rigorously in future work.

Third, we observe the effect of each component on the accuracy by deactivating the importance map or drop mask, respectively. The lower part of Table \ref{tab:tab1} summarizes the experimental results. From this, we can confirm that applying the drop mask and the importance map at the same time has better localization accuracy than applying only one of them. This supports the argument that the drop mask and importance map are not mutually exclusive.

\begin{table}[]
\small
\centering
\begin{tabular}{@{}lccc@{}}
\toprule
\begin{tabular}[c]{@{}l@{}}Applied\\ feature map\end{tabular} & \begin{tabular}[c]{@{}c@{}}GT-Known\\ Acc (\%)\end{tabular} & \begin{tabular}[c]{@{}c@{}}Top-1\\ Clas (\%)\end{tabular} & \begin{tabular}[c]{@{}c@{}}Top-1\\ Loc (\%)\end{tabular} \\ \midrule
\multicolumn{1}{l|}{N/A}                                 & 51.09                                                       & \multicolumn{1}{c|}{67.55}                                & 34.41                                                    \\ \midrule
\multicolumn{1}{l|}{\emph{conv 5-3}}                        & 57.99                                                       & \multicolumn{1}{c|}{\textit{68.95}}                       & 41.73                                                    \\
\multicolumn{1}{l|}{\emph{+ pool4}}                         & 68.22                                                       & \multicolumn{1}{c|}{67.17}                                & 48.02                                                    \\ 
\multicolumn{1}{l|}{\emph{+ pool3}}                         & \textbf{75.41}                                              & \multicolumn{1}{c|}{65.27}                                & \textbf{52.36}                                           \\ 
\multicolumn{1}{l|}{\emph{+ pool2}}                         & 71.85                                                       & \multicolumn{1}{c|}{63.76}                                & 48.46                                                    \\
\multicolumn{1}{l|}{\emph{+ pool1}}                         & 74.78                                                       & \multicolumn{1}{c|}{62.25}                                & 49.69                                                    \\ \bottomrule
\end{tabular}\vspace{2mm}
\caption{Effects in accuracy upon the choice of the feature maps to employ ADL. Bold text refers the best localization accuracy, while \textit{italic text} refers the best classification accuracy.}
\label{tab:tab2}
\end{table}
\begin{table*}[]
\centering
\small
\begin{tabular}{@{}lcccccccc@{}}
\toprule
\multicolumn{1}{c}{\multirow{2}{*}{Method}} & \multirow{2}{*}{Backbone} & \multirow{2}{*}{\begin{tabular}[c]{@{}c@{}}\# of \\ Params\\ (Mb)\end{tabular}} & \multicolumn{2}{c}{Overheads} & \multicolumn{2}{c}{CUB-200-2011}                                                                                     & \multicolumn{2}{c}{ImageNet-1k}                                                                                      \\ \cmidrule(l){4-9} 
\multicolumn{1}{c}{}                        &                           &                                                                                 & parameter (\%)  & computation (\%)  & \begin{tabular}[c]{@{}c@{}}Top-1\\ Loc (\%)\end{tabular} & \begin{tabular}[c]{@{}c@{}}Top-1\\ Clas (\%)\end{tabular} & \begin{tabular}[c]{@{}c@{}}Top-1\\ Loc (\%)\end{tabular} & \begin{tabular}[c]{@{}c@{}}Top-1\\ Clas (\%)\end{tabular} \\ \midrule
CAM                                         & VGG-GAP \cite{simonyan2014very, zhou2016learning}                   & 78                                                                              & 0            & 0             & 34.41                                                    & 67.55                                                     & \enspace42.80\mbox{*}                                                    & \enspace66.60\mbox{*}                                                     \\

ACoL                                        & VGG-GAP \cite{simonyan2014very, zhou2016learning}                   & 181                                                                             & 132.05       & 37.63         & \enspace45.92\mbox{*}                                                    & \enspace71.90\mbox{*}                                                     & \textbf{\enspace45.83\mbox{*}}                                                    & \enspace67.50\mbox{*}                                                         \\
ADL                                         & VGG-GAP \cite{simonyan2014very, zhou2016learning}                   & 78                                                                              & 0            & 0.00             & \textbf{52.36}                                                    & 65.27                                                     & 44.92                                                    & 69.48                                                     \\ \midrule
CAM                                         & MobileNetV1 \cite{howard2017mobilenets}              & 16                                                                              & 0            & 0             & 43.70                                                    & 71.94                                                     & 41.66                                                    & 68.38                                                     \\
HaS-32                                         & MobileNetV1 \cite{howard2017mobilenets}              & 16                                                                              & 0            & 0             & 44.67                                                    & 66.64                                                     & 41.87                                                    & 67.48                                                     \\
ADL                                         & MobileNetV1 \cite{howard2017mobilenets}               & 16                                                                              & 0            & 0.00             & \textbf{47.74}                                                    & 70.43                                                     & \textbf{43.01}                                                    & 67.77                                                     \\ \midrule
CAM                                         & ResNet50-SE \cite{he2016deep, hu2018squeeze}               & 107                                                                             & 0            & 0             & 42.72                                                    & 80.65                                                     & 46.19                                                    & 76.56                                                     \\

ADL                                         & ResNet50-SE \cite{he2016deep, hu2018squeeze}               & 107                                                                             & 0            & 0.00             & \textbf{\underline{62.29}}                                                    & 80.34                                                     & \textbf{48.53}                                                    & 75.85                                                     \\ \midrule
CAM                                         & InceptionV3 \cite{szegedy2016rethinking, zhang2018self}                & 101                                                                             & 0            & 0             & \enspace43.67\mbox{*}                                                    & -                                                         & \enspace46.29\mbox{*}                                                    & -                                                         \\

SPG                                         & InceptionV3 \cite{szegedy2016rethinking, zhang2018self}                 & 146                                                                             & 44.55        & 30.05         & \enspace46.64\mbox{*}                                                    & -                                                         & \textbf{\enspace48.60\mbox{*}}                                                    & -                                                         \\
ADL                                         & InceptionV3 \cite{szegedy2016rethinking, zhang2018self}                 & 101                                                                             & 0            & 0.00             & \textbf{53.04}                                                    & 74.55                                                     & \textbf{\underline{48.71}}                                                    & 72.83                                                     \\ \bottomrule
\end{tabular}\vspace{2mm}
\caption{Quantitative evaluation results on CUB-200-2011 and ImageNet-1k. Bold text refers the best localization accuracy for each backbone network. We also underline the best score in each dataset. Overheads are computed based upon their backbone networks. The accuracy with asterisk\mbox{*} indicates that the score is from the original paper. We leave some \textit{Top-1 Clas} scores blank, because they are not reported in the original paper \cite{zhang2018self}. For reproducing baseline methods, we use hyperparameters suggested by their original papers \cite{zhou2016learning, singh2017hide}. Also, we train and test HaS and ADL under the same setting for a fair comparison.}
\label{tab:tab3}\vspace{-3mm}

\end{table*}

When the importance map is applied alone, the classification accuracy increases but the localization accuracy decreases. We believe that this is because the classifier focuses more on the most discriminative part, guided by the importance map. This result supports our argument that the proposed lightweight attention method is effective to improve the classification accuracy. On the other hand, when drop mask is applied alone, the localization accuracy increases but the classification accuracy decreases. We believe that this is because the model utilizes less discriminative parts for classification, guided by the drop mask. These results also support the observation that the accuracy of localization and classification are in a trade-off relationship when applying the drop mask \cite{singh2017hide}.

Lastly, we investigate the effects in accuracy upon the choice of feature maps where ADLs are employed and report the results in Table \ref{tab:tab2}. From these results, we can see that applying ADLs to additional convolutional feature maps further increases the localization accuracy. We find that the ADL can improve both localization and classification accuracy. However, the best localization accuracy can be achieved by sacrificing the classification accuracy. In addition, when the ADLs are applied to lower-level feature maps such as \emph{pool2} and \emph{pool1}, the localization accuracy rather decreases. We believe that this is because the lower-level feature maps include general features that are not related to the target class. Consequently, the most discriminative part cannot be effectively eliminated in lower-level feature maps using ADL.

\subsection{Comparison with State-of-the-art Methods}

We compare the proposed method with various recent WSOL techniques including the state-of-the-art: CAM \cite{zhou2016learning}, HaS \cite{singh2017hide}, ACoL \cite{zhang2018adversarial}, SPG \cite{zhang2018self}. We report the accuracy of ACoL and SPG from their original paper. Meanwhile, we train the backbone networks using the same pre-processing method used in ACoL and SPG. Then, HaS or ADL are applied on the backbone networks. Considering the accuracy of vanilla model as baseline, we evaluate the accuracy gain of HaS and ADL, respectively. Please note that ACoL and SPG are the current state-of-the-art techniques for WSOL. In addition, among the techniques without parameter overheads, HaS performs the best.

Figure \ref{fig:qualitative} visualizes the localization results on CUB-200-2011 and ImageNet-1k dataset for qualitative evaluation. From the results, we consistently observe that model with ADL captures the less discriminative parts better than vanilla model. For example, as seen from the left-most sample in the Figure \ref{fig:qualitative}, the heatmap and bounding box extracted from vanilla model only highlight the face of birds. Contrarily, the model with ADL covers not only the face, but also the entire part of the bird, from head to wing. In addition, from the right-most sample in the Figure \ref{fig:qualitative}, the vanilla model focuses only on the cylinder of \textit{revolver}, whereas the model with ADL localizes the entire frame of the \textit{revolver}.

Next, the quantitative evaluation results on CUB-200-2011 and ImageNet-1k datasets are summarized in Table \ref{tab:tab3}. To compare the computing resources required by each technique, we have described the number of parameters and both computation and parameter overheads along with \textit{Top-1 Loc} and \textit{Top-1 Clas}. ADL has no parameter overheads, and the computation overheads are nearly zero (\textit{e.g.}, 0.003\% in ResNet50-SE) upon the backbone network. The proposed method is much more efficient than the existing state-of-the-art techniques, ACoL and SPG, in terms of both parameter and computation overheads. 

We push further to maximize the efficiency of WSOL by employing MobileNetV1 \cite{howard2017mobilenets} as a backbone network. Due to the lightweight nature of MobileNetV1, it is inappropriate to employ ACoL or SPG which requires huge additional computing resources. On the other hand, ADL and HaS can be successfully employed despite a limited amount of resources. From the experimental results, we can observe that the accuracy gain of the proposed method is better than that of HaS. In addition, HaS has reduced classification accuracy against the baseline. This is caused by the trade-off relationship between localization and classification accuracy discussed in Section \ref{sec:adl}. Fortunately, the importance map of ADL can subside such a drawback by increasing the classification power. Consequently, the classification accuracy degradation of the ADL is not as significant as that of HaS. 

In addition to its high efficiency, the proposed method achieves a new state-of-the-art localization accuracy on CUB-200-2011 dataset. When ResNet50-SE is employed as a backbone, the proposed method improves the localization accuracy by more than 15 percentage points over the state-of-the-art accuracy \cite{zhang2018adversarial, zhang2018self}. Please note that the number of parameters of ResNet50-SE with ADL is much smaller than that of ACoL and SPG. This achievement is quite impressive, considering that recent techniques are competing with the accuracy by 2-3 percentage points difference. Also, when the other three backbone networks are employed, the proposed method achieves better localization accuracy than the existing state-of-the-art techniques. 

In the ImageNet-1k experiments, when VGG-GAP is used as a backbone, the accuracy of ADL is better than that of CAM, but slightly lower than that of ACoL. However, when ResNet50-SE is used as a backbone, localization accuracy of ADL is better than that of ACoL and comparable with that of SPG even though the required computing resources are much lower. In addition, when InceptionV3 is used as a backbone, comparable accuracy (0.11 percentage point difference) to SPG is achieved. In summary, we achieve new state-of-the-art accuracy on CUB-200-2011 dataset; on ImageNet-1k dataset, ADL achieves comparable accuracy with the current state-of-the-art technique \cite{zhang2018self} despite its superior efficiency.

\vspace{3mm}
\noindent \textbf{Discussion.} We verified the proposed method on a single-object detection task, following the current state-of-the-art methods \cite{zhang2018adversarial, zhang2018self}. However, it should be noted that the proposed method can be also used to improve the weakly supervised semantic segmentation accuracy. The classifier with ADL is the same as its vanilla version during testing, thus it can be easily combined with the weakly supervised semantic segmentation framework, such as \cite{kolesnikov2016seed, oh2017exploiting}.

\begin{figure}[t!]
    \noindent
    \begin{center}
        \subfigure[Input]{\includegraphics[width=0.30\columnwidth]{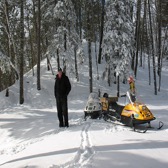}}\quad
        \subfigure[CAM]{\includegraphics[width=0.30\columnwidth]{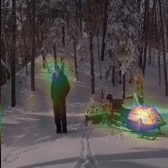}}\quad
        \subfigure[ADL (Ours)]{\includegraphics[width=0.30\columnwidth]{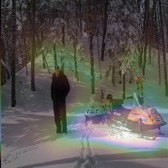}}
    \end{center}
    
    \caption{The failure case on ImageNet-1k experiments. The target class is \emph{snowmobile}. The model with ADL learns the less discriminative region which is not included in object. Specifically, the model captures not only the \textit{snowmobile}, but also \textit{snow} and \textit{tree}.}\vspace{-1mm}
\label{fig:fail}
\end{figure}

Next, to analyze the substantial difference in our accuracy gain between two datasets, we investigate our failure examples from ImageNet-1k experiments. From the failure case, we observe that the classifier extracts the discriminative features from the background which appears frequently with the target object. Figure \ref{fig:fail} illustrates such examples. In the case of the \emph{snowmobile} class, the target object often co-occurs with \textit{snow}. The vanilla model only focuses on the \textit{snowmobile}, while the model with ADL learns not only the \textit{snowmobile}, but also the \textit{snow} and \textit{tree}. This is because the background frequently appearing with the object might be the less discriminative region. 

ImageNet-1k includes a wide variety of classes where specific types of background co-occur with the target object. In this case, the background has a certain level of discriminative power. Therefore, the model is likely to learn the background features when the most discriminative part is dropped. 
Meanwhile, since all classes of CUB-200-2011 belong to birds, similar backgrounds appear regardless of the classes (\textit{e.g}. sky, tree). In other words, the background of this dataset is nearly independent of classes, thus the background is not a discriminative region \cite{zhang2016picking}. As a result, the model does not learn the features from the background although the most discriminative part is hidden. 

This explains the gap of our accuracy gain for two datasets; ADL has remarkable performance to induce the classifier to learn the less discriminative parts, as supported by CUB-200-2011 evaluations. We believe that this problem might be critical for all WSOL methods inducing the classifier to learn the less discriminative part. Currently, it seems non-trivial to solve this problem, thus we will address this issue in future work. Lastly, we note that the gap is not caused by the scale of dataset because ADL rarely fails for ImageNet-1k classes sharing similar background statistics (\textit{e.g.}, various breeds of dogs). 

\section{Conclusion}
We presented an \emph{Attention-based Dropout Layer} (ADL), a novel weakly supervised object localization method that induces the CNN classifier to learn entire extent of the object. The proposed method is much more efficient and lightweight than existing state-of-the-art methods. In addition, the proposed method has achieved excellent performance; new state-of-the-art accuracy on CUB-200-2011, and comparable accuracy with current state-of-the-arts on ImageNet-1k. We also demonstrate that the proposed method can be easily applied to various CNN classifiers to improve the localization accuracy. For the future work, we will analyze the regularization effect of the \emph{drop mask}. In addition, we will address the problem that the model learns the less discriminative region from outside of the object. 

\section*{Acknowledgement}
\small{
This research was supported by the Basic Science Research Program through the National Research Foundation of Korea (NRF) funded by the MSIP (NRF-2019R1A2C2006123), 
and the MIST (Ministry of Science and ICT), Korea, under the ``ICT Consilience Creative Program'' (IITP-2018-2017-0-01015) supervised by the IITP (Institute of Information \& Communications Technology Planning \& Evaluation). 
This work was also supported by ICT R\&D program of MSIP/IITP [R7124-16-0004, Development of Intelligent Interaction Technology Based on Context Awareness and Human Intention Understanding]. }

\begin{figure*}[h]
\centering
\includegraphics[width=2.0\columnwidth]{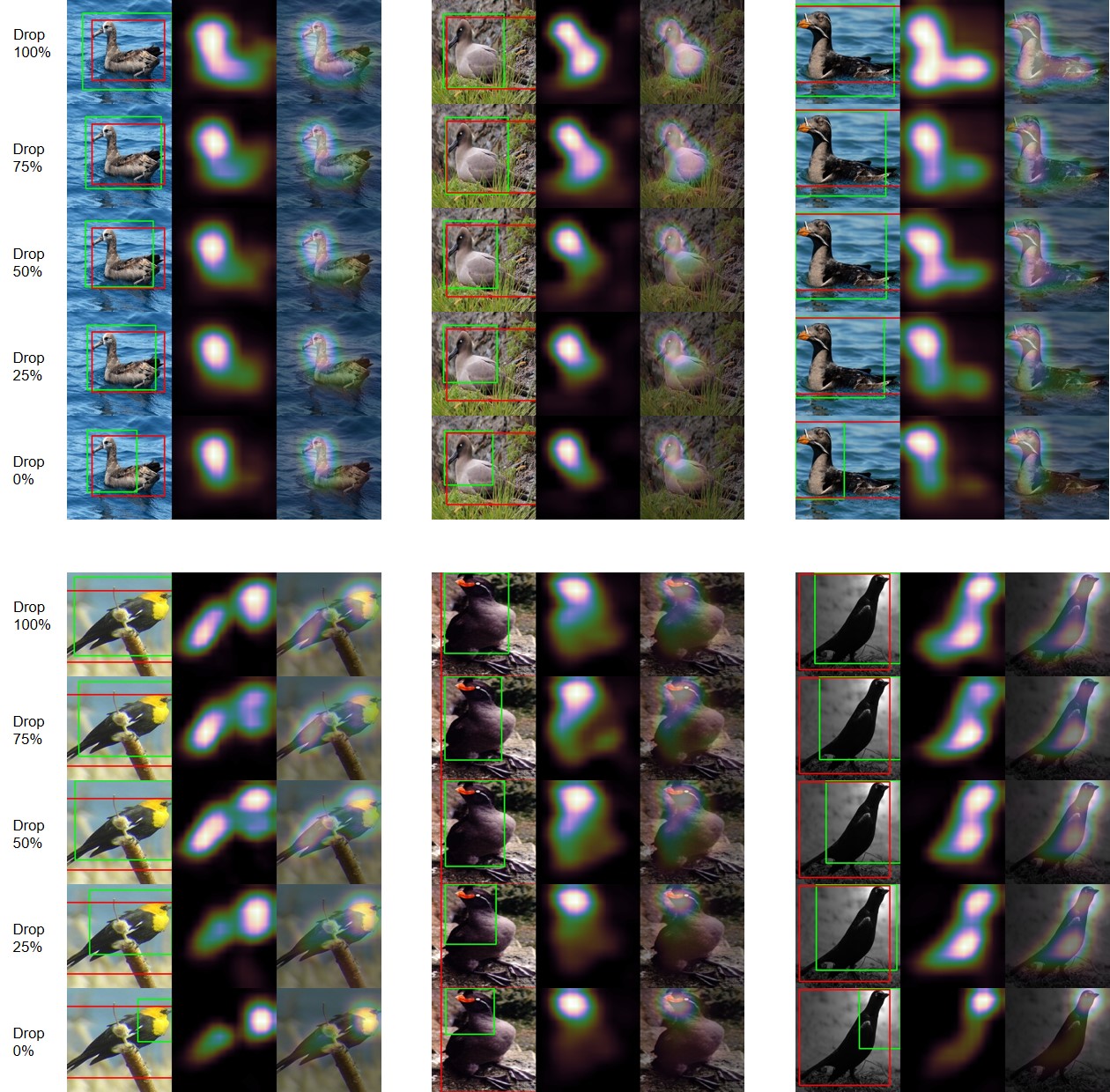}
\caption{Qualitative evaluation results corresponding to Table \ref{tab:tab1}. As the drop rate decreases, we can see that the model focuses only on the most discriminative part of the object.}
\label{fig:qual1}
\end{figure*}
\begin{figure*}[h]
\centering
\includegraphics[width=2.0\columnwidth]{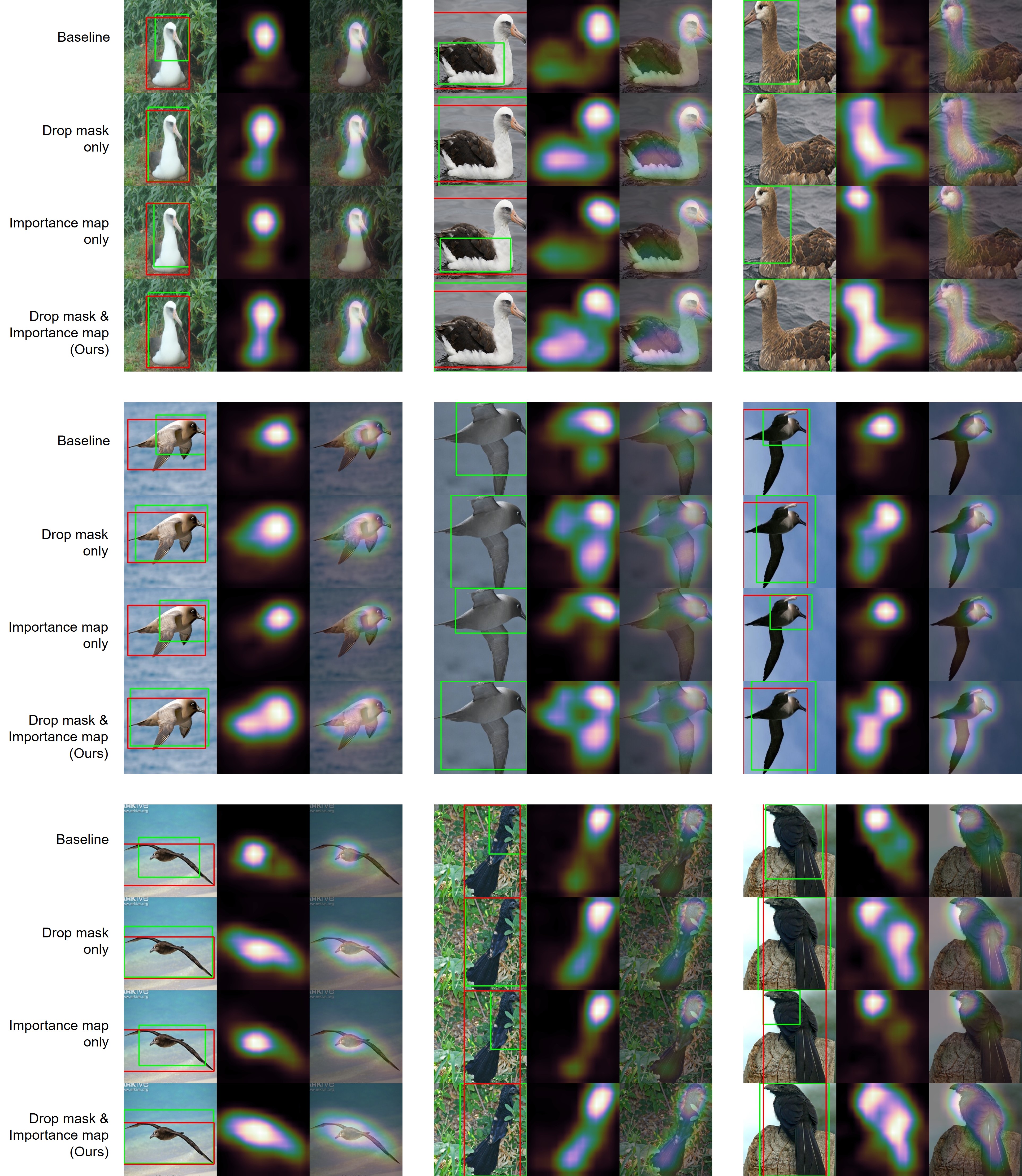}
\caption{Qualitative evaluation results corresponding to Table \ref{tab:tab1}. When only an importance map is applied, the model focuses only on the most discriminative part object. On the other hand, when the drop mask is applied, the model can localize the entire extent of object. We note that the best localization result can be obtained using both drop mask and importance map.}
\label{fig:qual2}
\end{figure*}
\begin{figure*}[h]
\centering
\includegraphics[width=2.0\columnwidth]{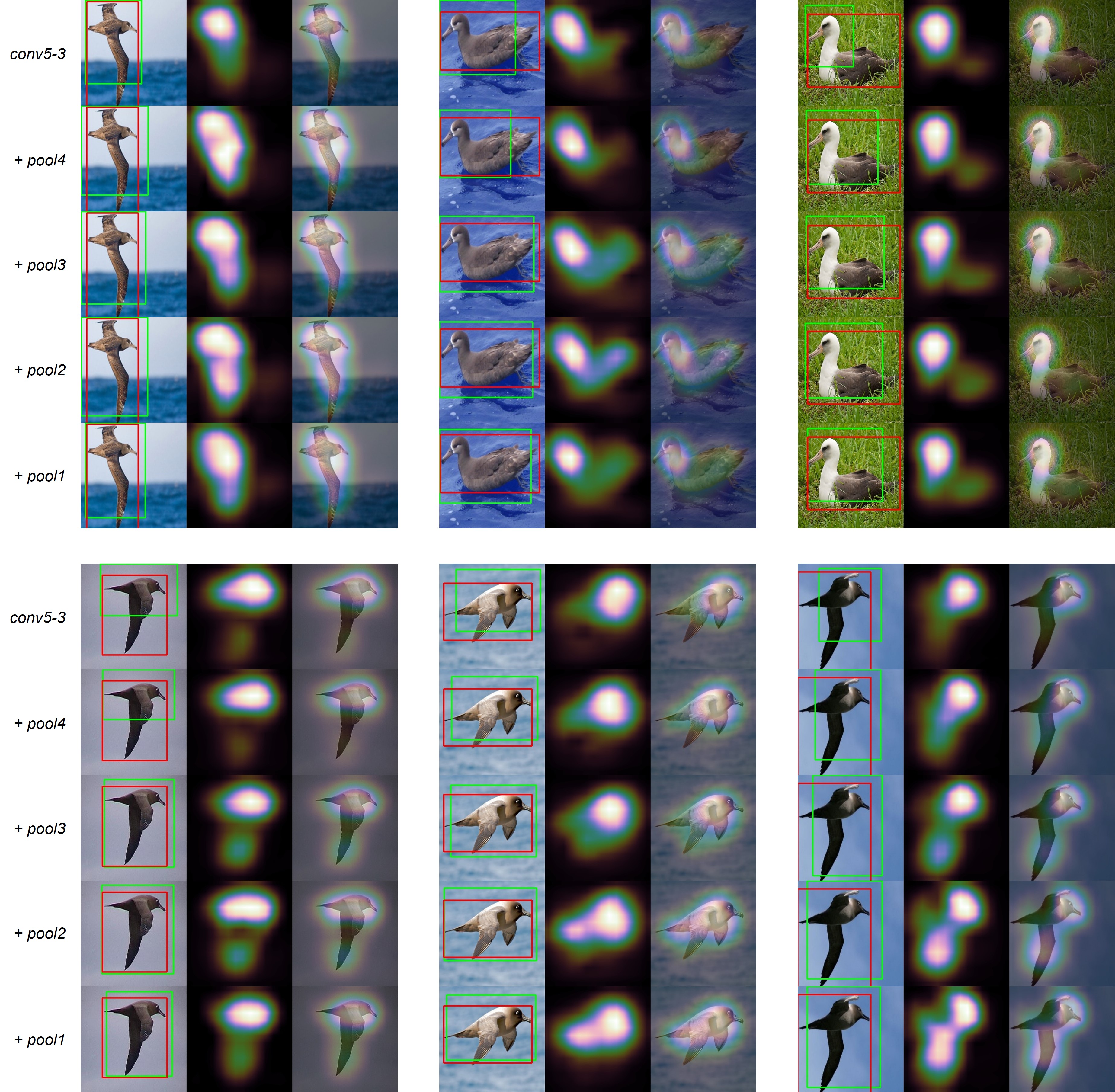}
\caption{Qualitative evaluation results corresponding to Table \ref{tab:tab2}. We can see that the localization accuracy increases when the ADLs are applied on multiple feature maps.}
\label{fig:qual3}
\end{figure*}
\begin{figure*}[h]
\centering
\includegraphics[width=2.0\columnwidth]{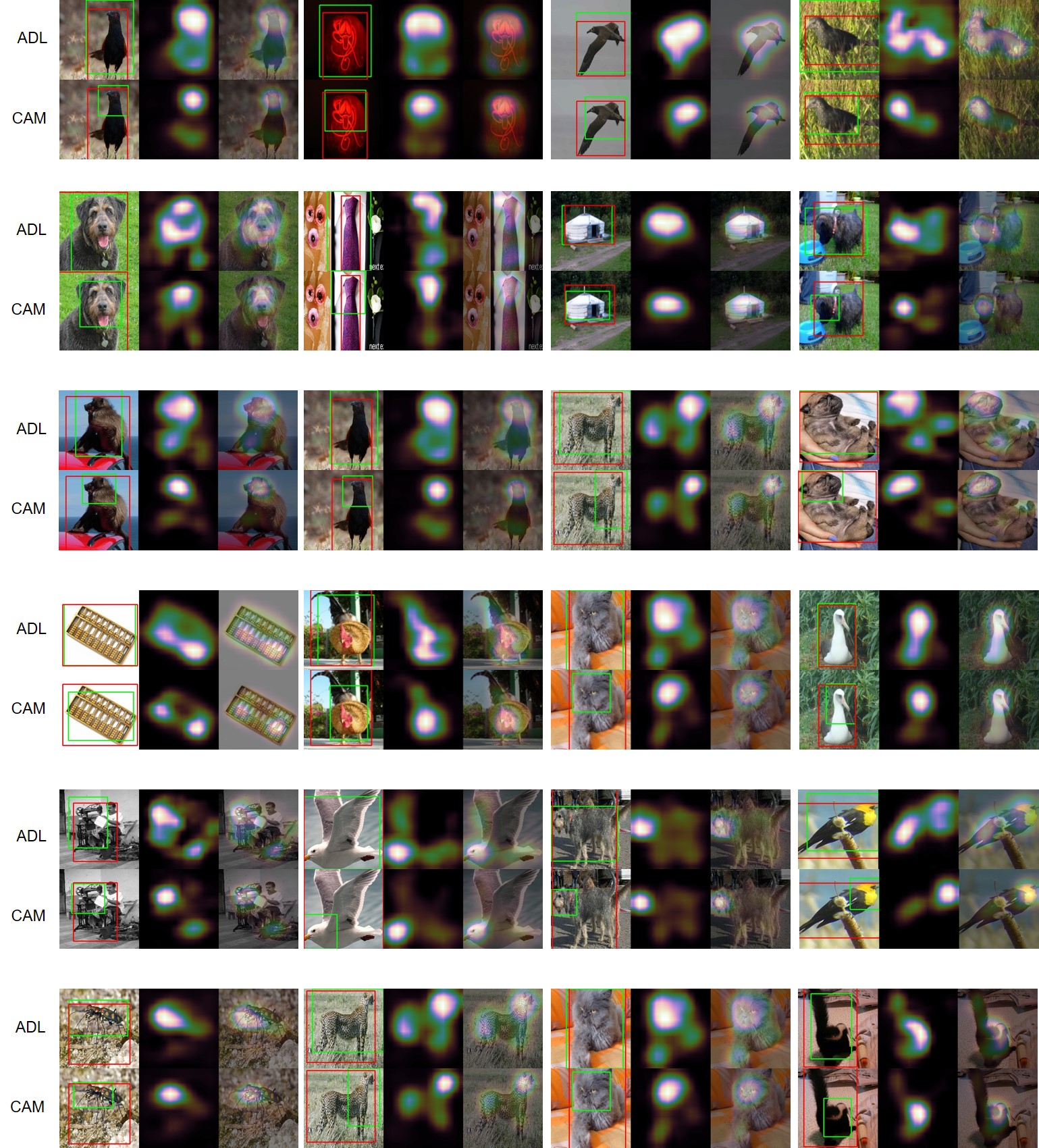}
\caption{More qualitative evaluation results. The left image in each figure is input image. The red bounding box is ground truth, while the green bounding box is estimates. The middle image is heatmap and the right image shows the overlap between the input image and the heatmap. We also compared our method and the vanilla model side by side.}
\label{fig:qual4_app}
\end{figure*}

\end{document}